\documentclass{bmvc2k}

\usepackage{graphicx}
\usepackage{multicol}
\usepackage{xspace}
\usepackage{amssymb}
\usepackage{adjustbox} % to squeeze the table into the page
\usepackage{amsmath}
\usepackage{mathtools}

\usepackage[activate={true,nocompatibility},final,tracking=true,kerning=true,spacing=true,factor=1100,stretch=10,shrink=9]{microtype}

\DeclareMathOperator*{\argmax}{arg\,max}

\newcommand{\squeezeup}{\vspace{-2mm}}

\newcommand{\eg}{e.\,g.\xspace}

\newcommand{\tick}{\checkmark}
\title{Semantic Estimation of 3D Body Shape and Pose using Minimal Cameras}

\addauthor{Andrew Gilbert}{https://www.surrey.ac.uk/people/andrew-gilbert}{1}
\addauthor{Matt Trumble}{}{1}
\addauthor{Adrian Hilton}{https://www.surrey.ac.uk/people/adrian-hilton}{1}
\addauthor{John Collomosse}{http://personal.ee.surrey.ac.uk/Personal/J.Collomosse/index.php}{12}

\addinstitution{
 Centre for Vision Speech and Signal Processing,\\
 University of Surrey,\\
 Guildford, \\
 UK\\
}
\addinstitution{
 Creative Intelligence Lab,\\
 Adobe Research, USA\\
}

\runninghead{Gilbert \emph{et al}}{Semantic Estimation of 3D Body Shape and Pose}

\def\eg{\emph{e.g}\bmvaOneDot}

\begin{document}

\maketitle

\begin{abstract}
We aim to simultaneously estimate the 3D articulated pose and high fidelity volumetric occupancy of human performance, from multiple viewpoint video (MVV) with as few as two views. We use a multi-channel symmetric 3D convolutional encoder-decoder with a dual loss to enforce the learning of a latent embedding that enables inference of skeletal joint positions and a  volumetric reconstruction of the performance. The inference is regularised via a prior learned over a dataset of view-ablated multi-view video footage of a wide range of subjects and actions, and show this to generalise well across unseen subjects and actions.   We demonstrate improved reconstruction accuracy and lower pose estimation error relative to prior work on two MVV performance capture datasets: Human 3.6M and TotalCapture.
\end{abstract}
\squeezeup
\squeezeup
%%%%%%%%% BODY TEXT
\begin{figure}[htp]
\centering
\includegraphics[width=1\textwidth]{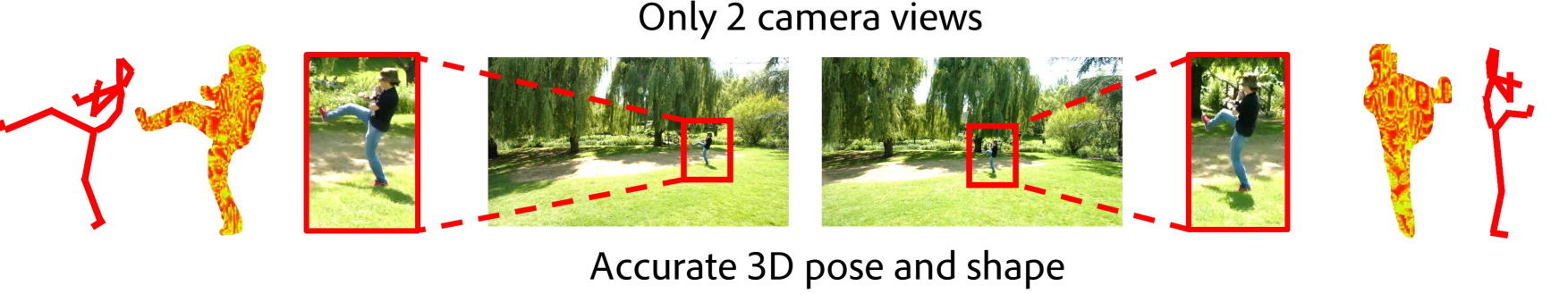}
\end{figure}
\section{Introduction}
%Generic overview of human motion
Human performance capture is used extensively within biomechanics and the creative industries.  Commercial approaches are typically constrained to skeletal joint estimation in the presence of subject-worn markers captured from multiple viewpoints by specialised (e.g. infra-red) cameras.
In this paper, we present a method for video-based performance capture, able to estimate both 3D skeletal pose and shape (volumetric occupancy) of a subject accurately from multiple-viewpoint video (MVV). Uniquely, we do so without a parametric shape model (\eg SMPL \cite{loper2015SMPL}), and without the need for worn markers or sensors~\cite{PonsECCV18,trumble_total_2017}, nor a large camera count \cite{collet2015MSFVV}.  Our approach considers MVV with as few as two wide baseline cameras, motivated by real-world scenarios that may constrain the on-set deployment of large numbers of witness camera views due to limitations on camera cost or placement (\eg security or sports events).

%This work proposes to use a deeply learnt prior before recovering a high fidelity geometric proxy of the subject from a coarse input, along with their skeletal pose. Regularised by a generative adversarial network to improve the realism of the resultant geometric proxy.
%full markers intrusive and one camera gives not great results but requiring many cameras at run time is also bad
%Motion capture (mo-cap) technology has its origins in biomechanics, where the analysis of human performance data has been used to inform diagnosis and training strategy. However, the past decade has seen applications of mo-cap broaden to include performance capture, e.g. to add realism and reduce the cost of character animation in the creative industries. However, existing commercial solutions (e.g. Vicon, OptiTrack) are typically reliant upon specialist camera equipment such as active or retro-reflective infra-red markers, stereo-triangulation depth sensors and time-of-flight cameras. While research approaches are highly effective in 2D pose estimation~\cite{wei2016cpm} or through the inclusion of addition sensors such as IMUs~\cite{PonsECCV18,trumble_total_2017}, or the requirement of many cameras~\cite{trumble:eccv:2018,pavlakos2017volumetricCVPR}. These place restrictions on the capture environment, such as prohibiting or limiting outdoor shoots, as well as restricting the size of the ‘capture volume’. 

%Ta da! the answer is minimal cameras with a dual loss using a generative network to constrain the appearance of the PVH
Our technical contribution is to learn a generative model that accepts a coarse poor quality volumetric proxy formed from a low number of wide baseline camera views of a subject. In a single inference step, we estimate both the skeletal joint positions (pose) and refine a higher fidelity volumetric reconstruction from the rough proxy (occupancy).

Our architecture is a volumetric encoder-decoder convolutional neural network (CNN) in which the latent bottleneck is partially constrained to estimate the 3D skeletal pose and partially unimpeded to enhance the fidelity of volumetric reconstructions derived from just a few wide-baseline camera viewpoints. A joint loss between both outputs is used within a generative adversarial network to ensure the refinement of the volumetric solution to enable it to be perceptually indistinguishable from real high-fidelity reconstructions restoring fine detail such as hands and legs. Unlike prior work that has explored volumetric encoder-decoder networks for pose \cite{trumble:eccv:2018} or for content up-scaling \cite{gilbert2018volumetric}, we leverage use of 2D semantic detections to supplement the background occupancy volumetric proxy. The encoder-decoder network serves to learn a prior for human shape, regularised by a generative adversarial network (GAN) loss that ensures realism in the output high-fidelity volumetric reconstruction output and enabling both the pose estimation and reconstruction to be learnt from a minimal set of camera views. The work by Trumble \emph{et al}~\cite{trumble:eccv:2018}  inspires this work, however with significantly improved performance through the introduction of several notable novelties; the inclusion of semantic labels as well as occupancy probabilities in the voxels that make the PVH.  The incorporation of a GAN discriminator on the output volume and the extension of the bottleneck of the encoder-decoder with additional latent features besides the body joint coordinates. We demonstrate SOTA results and several ablation studies in the paper which show the value of these contributions.

\squeezeup
\squeezeup
%%%%%%%%%%%%%%%%%%%%%%%%%%%%%%%%%%%%%%%%%%%%%%%%%%%%%%%%%%%%%%%%
\section{Related Work}
Our work is inspired by contemporary super-resolution (SR) algorithms that apply learned priors to enhance visual detail in images, volumetric performance capture or reconstruction and human pose estimation (HPE).

{\bf Super-resolution:} Classical image restoration / SR approaches combine multiple data sources (\eg images \cite{Fattal2007}, or self-similar patches \cite{Glasner2009,Zhu2014}) under regularization \eg total variation \cite{tvexample}.  Convolutional neural network (CNN) autoencoders have been applied to image \cite{Xie2012,Wang2015,Dong2016} and video-upscaling \cite{Shi2016}. Volumetric SR has been explored for microscopy~\cite{Abrahamsson2017}, and for multi-spectral sensing \cite{Atalay2017}. Recently SR for volumetric performance capture was explored using encoder-decoder networks \cite{gilbert2018volumetric}.

{\bf Volumetric Performance Reconstruction:} Volumetric performance capture pipelines typically use multiple wide baseline viewpoints \cite{starck2009FVVR,casas2014rwvc} arranged around the capture volume.  More recently, data driven machine learning approaches~\cite{VarolECCV18,JacksonECCV18,ZhengICCV19} have demonstrated improve reconstruction from a single camera. Varol \emph{et al}~\cite{VarolECCV18} use a neural network for direct inference of volumetric body shape from a single image. While Jackson \emph{et al}~\cite{JacksonECCV18} directly regress the volumetric representation of the 3D geometry using a standard, spatial, CNN architecture, and Zheng \emph{et al}~\cite{ZhengICCV19} also uses the parametric representation of the SMPL body model~\cite{loper2015SMPL} fusing different scales of image features into the 3D space through volumetric feature transformation, to recover accurate surface geometry. 

{\bf Human Performance Estimation:}  There are two distinct categories of HPE; bottom-up data-driven and top-down fitting a model. In general, top-down 2D pose estimation fits a previously defined articulated limb model to data incorporating kinematics into the optimisation to bias toward possible configurations. The model can be user-defined or learnt through a data defined model such as the SMPL Body Model~\cite{loper2015SMPL}. Spatio-temporal tracking of pictorial structures is applied to HPE in~\cite{lan04}, and~\cite{andriluka09} explored the fusion of pictorial structures with Ada-Boost shape classification. Malleson \emph{et al}~\cite{Malleson3DV17} used IMUs with a full kinematic solve to adequately estimate 3D pose both indoor and outdoor. Recently, the SMPL model has been employed by several pose estimation techniques with IMUs~\cite{SIP2017EG,PonsECCV18} and 2D images~\cite{TanSMPLY2d3DBMVC17,Huang3DV}.

%Marcard~\cite{SIP2017EG} fitted IMU measurements to the data defined SMPL model providing pose estimation without visual data. Tan~\cite{TanSMPLY2d3DBMVC17} employs the SMPL model to estimate the 3D pose from 2D images in a decoder/encoder framework. At the same time, Huang~\cite{Huang3DV} combines the SMPL body model with 2D joint estimates to reinforce and improve the 3D pose. More recently Marcard~\cite{PonsECCV18} used many IMUs and a single camera to recover 3D pose with the constraint of the SMPL model.

Bottom-up pose estimation is driven by image parsing to isolate components, Srinivasan \emph{et al}~\cite{srinivasan07} used graph-cuts to parse a subset of salient shapes from an image and group these into a model of a person. 
%However, the approach is sensitive to clutter which interferes with the segmentation. 
%Mori~\cite{mori04} identified joint positions, using scale and symmetry constraints between a 2D query image training images. 
Ren  \emph{et al}~\cite{ren05} recursively splits Canny edge contours into segments, classifying each as a putative body part using cues such as parallelism. Ren~\cite{ren12} also used Bag of Visual Words for implicit pose estimation as part of a pose similarity system for dance video retrieval. 
%More recently, studies have begun to leverage the power of convolutional neural networks, following in the wake of the eye-opening results of Krizhevsky ~\cite{Krizhevsky2012} on image recognition. 
In DeepPose, Toshev~\cite{Toshev2014} used a cascade of convolutional neural networks to estimate 2D pose in images. 
%Descriptors learnt by a CNN have also been used in 2D pose estimation from very low-resolution images \cite{Park2015}. 
Elhayek \emph{et al}~\cite{elhayek_efficient_2015} used MVV with a Convnet to produce 2D pose estimations while Rhodin \emph{et al}~\cite{rhodin2016general} minimised the edge energy inspired by volume ray casting to deduce the 3D pose. 
%More recently given the success and accuracy of 2D joint estimation~\cite{cao2016realtimeCPM}, several works lift 2D detections to 3D using learning or geometric reasoning, aiming to recover the missing depth dimension in the images. Sanzari~\cite{sanzari2016bayesianH36m} estimates the location of 2D joints, before predicting 3D pose using appearance and probable 3D pose of the discovered parts with a hierarchical Bayesian model. 
 %While Zhou~\cite{zhou2016sparsenessH36m} integrates 2D, 3D, and temporal information to account for uncertainties in the data. The challenge of estimating 3D human pose from MVV is currently less explored, generally casting 3D pose estimation as a coordinate regression task, with the target output being the spatial $x,y,z$ coordinates of a joint to a known root node such as the pelvis. 
 Trumble \emph{et al}~\cite{TrumbleCVMP2DConvNet} used a flattened MVV based spherical histogram with a 2D convnet to estimate pose. While Pavlakos  \emph{et al}~\cite{pavlakos2017volumetricCVPR} used a simple volumetric representation in a 3D convnet for pose estimation and Wei \emph{et al}~\cite{wei2016cpm} performed related work in aligning pairs of joints to estimate 3D human pose. 
 Since detecting pose for each frame individually leads to incoherent and jittery predictions over a sequence, many approaches exploit temporal information~\cite{andriluka2014MPI2DPoseDataset,lin2017CVPRRPSM} often using LSTMs~\cite{hochreiter1997LSTM}. 
Trumble et al. \cite{trumble:eccv:2018} estimate 3D pose using the latent space of a volumetric encoder-decoder, but do not incorporate semantic information nor GAN constraint.

\squeezeup
\squeezeup

%%%%%%%%%%%%%%%%%%%%%%%%%%%%%%%%%%%%%%%%%%%%%%%%%%%%%%%%%%%%%%%%
\section{Joint minimal camera Pose and Volume reconstruction}
\begin{figure*}[t!]
\centering
\includegraphics[width=0.6\linewidth]{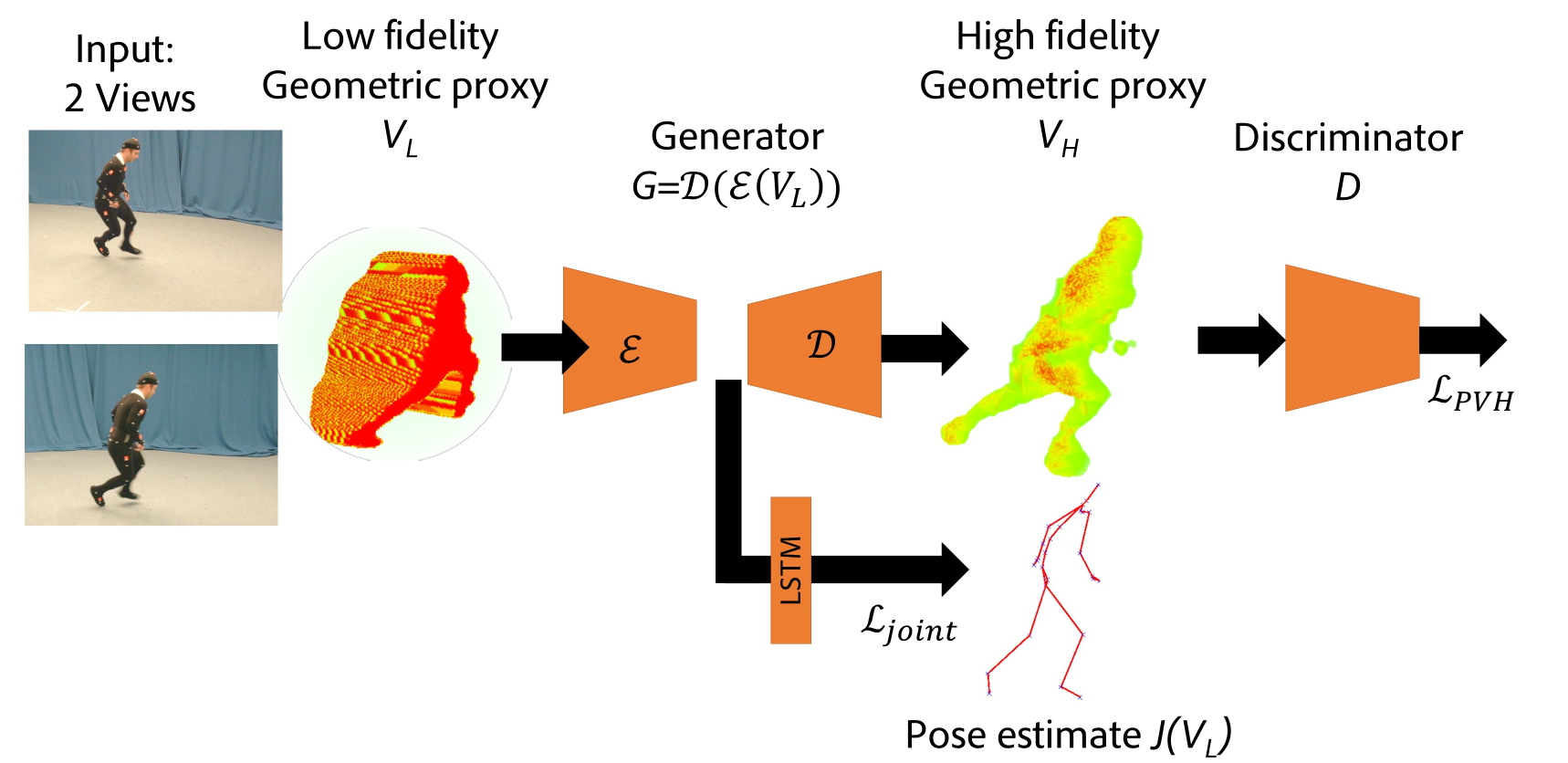}
   \caption{Network architecture. The input is a low fidelity geometric proxy ($V_L$) from two wide baseline camera views. This proxy is passed through a decoder-encoder to produce a 3D human pose estimate (joint angles $J(V_L)$) via the latent space and to output, a high-fidelity geometric proxy ($V_H$) regularised via discriminator (D).\label{fig:overview}
}
\end{figure*}
We present an overview of our process for simultaneously estimating pose and high fidelity occupancy in Figure~\ref{fig:overview}.  First, a pre-processing step~\cite{Grauman2003} reconstructs a coarse Probabilistic Visual Hull (PVH) proxy using a limited number of cameras (Sec.~\ref{sec:PVH}).  For each voxel, we encode a feature reflecting its occupancy and semantic label (e.g. joints) lifted from 2D.  This initial estimate (Sec.~\ref{sec:2DDectections}) typically contains phantom limbs and sub-volumes. Next, a 3D convolutional encoder-decoder (Sec.~\ref{sec:autoenc}) and generative adversarial network (GAN) (Sec.~\ref{sec:GAN}), learns a deep representation of body shape and the skeletal pose encoding with a dual loss. The feature representation of the PVH (akin to a low-fidelity image in super-resolution pipelines), is deeply encoded via a series of convolution layers, embedding the skeletal joint positions in a latent or hidden layer, concatenating the joint estimates with an additional unconstrained feature representation. This latent space enables non-linear mapping decoding to a high fidelity PVH, while the 3D joint estimations are fed to LSTM layers to enforce the temporal consistency of the 3D joints (Sec.~\ref{sec:TempConsistency}). 
%We also describe the data augmentation and methodology for training the 3D convolutional network (Sec.~\ref{sec:TCSetup}).
\squeezeup
\squeezeup

%%%%%%%%%%%%%%%%%%%%%%%%%%%%%%%%%%%%%%%%%%%%%%%%%%%%%%%%%%%%%%%%
\subsection{Visual Features }
\label{sec:2DDectections}
To estimate the pose, we propose to lift 2D visual features to form a 3D voxel features from two distinct modes created from RGB images of each camera view; a 2D foreground occupancy matte and 2D semantic joint detections. The probabilistic occupancy provides a low fidelity shape-based feature, relatively invariant to appearance and clothing, that complements a semantic contextual 2D joint estimate that provides internal feature description. To compute the matte, the difference between the current frame $I$ and a predefined clean plate $P$  approximates pixel occupancy. A thresholded $L2$ distance between the two images in the HSV colour domain provides a soft occupancy probability. 2D joint belief labels estimated through the approach of Wei~\cite{wei2016cpm,cao2017realtime} generate the 2D semantic joint detections, a multi-stage process that iteratively refines the 2D joint estimates based on both the input image and the previous stage’s returned pixel-wise belief map. At each stage $s$ and for each joint label $j$ the algorithm returns dense per pixel belief maps $m^{j}_{s}$, which provides the confidence of a joint centre for any given pixel $(x,y)$. 
\begin{eqnarray}
M(x,y) = \argmax_{j} m_{S}^{j}(x,y) \label{eq:beliefmap}
\end{eqnarray}
The per joint belief maps are maximised over the confidence of all possible joint labels to produce a single label per pixel image $M(x,y)$. 
%Fig.~\ref{fig:OpenPoseEx} illustrates the 2D occupancy and semantic joint labels for an example frame,  for this complex pose the occupancy shape is detected well, however, 
%There is often ambiguity over the 2D pose, due to uncertainties with left vs righthand limbs, However, by jointly using both occupancy and semantics to form voxel features in the PVH, accurate 3D pose estimates and shape proxy can be identified.
%\begin{figure}[htp]
%\centering
%\includegraphics[width=0.4\linewidth]{figures/OpenPoseEx.jpg}
%   \caption{An example of the foreground occupancy and 2D semantic labels %converted into probably for PVH construction.\label{fig:OpenPoseEx}
%}
%\end{figure}
%\squeezeup
\squeezeup

%%%%%%%%%%%%%%%%%%%%%%%%%%%%%%%%%%%%%%%%%%%%%%%%%%%%%%%%%%%%%%%%
\subsection{Volumetric Representation}
\label{sec:PVH}
%There exist several methods to estimate 3D pose; through multiple separate 2D views~\cite{pavlakos2017CVPRharvesting,TrumbleCVMP2DConvNet} that require many cameras in the scene or by inferring 3D from a single 2D view~\cite{tome2017liftingH36m,chen20163dCVPR,PonsECCV18}, which can fail for complex poses occluded by the single-camera view. However, by taking inspiration from super-resolution work, we propose a learn a generative approach that uses a minimal number of camera views and an inherent poor input to a learn a complex mapping to a multi-view 3D pose previously learnt from many camera views. Thus, learning to resolve complex ambiguities and occlusions present in individual 2D images. 

To construct our data representation consisting of a volume voxel, we use a multi-channel based probabilistic visual hull (PVH). 
We assume a capture volume  observed by a limited number $C$ of camera views $c=\left[ 1,..,C \right]$ for which extrinsic parameters $\{R_c, {COP}_c\}$ (camera orientation and focal point) and intrinsic parameters $\{f_c, o^x_c, o^y_c\}$ (focal length, and 2D optical centre) are known. An external process, (\eg a person tracker) isolates the bounding sub-volume $X_I \in \mathcal{V}$  corresponding to, and centred upon, a single subject, and which  is decimated into voxels $\mathbf{V_L}^i=\left[\begin{array}{ccc} v_x^i &v_y^i &v_z^i \end{array}\right]$ for $i=\left[1, \dots, |\mathbf{V_L}|\right]$; each voxel is $5 \mathrm{mm}^3$ in size. Each voxel $v^i \in \mathbf{V_L}$ projects to coordinates $(x[v^i],y[v^i])$ in each camera view $c$.

Then given an 2D image denoted as $I_c$, with $\Phi =\left[1, \dots, \phi\right] $ feature channels (from 2D occupancy/joints), point $(x_c,y_c)$ is the point within $I_c$ to which $\mathbf{V_L}^i$ projects in a given view:
\begin{eqnarray}
x[\mathbf{V_L}^i]&=&\frac{f_c v_x^i}{v_z^i}+o^x_c ~~~\mathrm{and} ~~~y[\mathbf{V_L}^i]=\frac{f_c v_y^i}{v_z^i}+o^y_c,\\
\left[
\begin{array}{ccc} v_x^i &v_y^i &v_z^i\end{array}\right] &=& {COP}_c - R_c^{-1} V_L^i. \label{eq:pvh3}
\end{eqnarray}
The likelihood of the voxel being part of the performer in a given view $c$ is: 
\begin{eqnarray}
p(\mathbf{V_L}^i | c) = I_c(x[\mathbf{V_L}^i],y[\mathbf{V_L}^i],\phi).  \label{eq:pvh1}
\end{eqnarray}
The overall probably of occupancy for a given voxel $p(\mathbf{V_L}^i,\phi)$ is:
\begin{eqnarray}
p(\mathbf{V_L}^i,\phi) = \prod_{i=1}^C 1/(1+e^{-p(\mathbf{V_L}^i|c)}). \label{eq:pvh4}
\end{eqnarray}
%We are then able compute $p(\mathbf{V_L}^i)$ for all voxels to create the PVH for volume $\mathbf{V_L}$.  
%\begin{eqnarray}
% \sum_{i \in V} \sum_{j \in \Phi} p(v_i,\phi_j) \label{eq:pvh5}
%\end{eqnarray}

%The fine-grained voxel occupancy approximation is then downsampled via a weighted Gaussian filter to the coarse input shape and size of the first layer in the convnet, 30x30x30, this roughly approximates with the same number of pixels as a $150x150$ 2D image, where each voxel approximates a 67x67x67mm volume in the real world. 

%%%%%%%%%%%%%%%%%%%%%%%%%%%%%%%%%%%%%%%%%%%%%%%%%%%%%%%%%%%%%%%%
\subsection{Dual Loss Convolutional Volumetric Network}
\label{sec:autoenc}
%At their simplest, an encoder-decoder neural network architecture learns an encoding from an input signal domain by training the network to reconstruct the input through a bottleneck layer of reduced dimensionality (the latent embedded space). A concatenation of the 3D pose estimates and a vector with no direct constraint. By using multiple 2D views, our result can generate a realistic 3D representation and pose of the human body, thus able to avoid ambiguities and occlusions present in independent, individual 2D images. 

%Need to check the XYZ1 for the CPM space?
We propose to learn a deep representation or output given an input tensor $\mathbf{V_L}$ where $\mathbf{V_L} \in \mathbb{R}^{X \times Y \times Z \times \phi}$, where each dimension encodes the probability of volume occupancy $p(X,Y,Z)$ derived from a PVH obtained using a low camera count (Eq.\ref{eq:pvh4}) from channels ($\phi$); foreground occupancy  and semantic 2D  joint  estimates.  We wish to train a deep representation to solve the prediction problem $\mathbf{V_H} = \mathcal{F}(\mathbf{V_L})$ for similarly encoded tensor $\mathbf{V_H} \in \mathbb{R}^{W \times H \times D \times \phi}$ derived from a higher fidelity PVH of identical dimension obtained using a  higher camera count. Where $W,H,D,\phi$ are the width, height, depth and channel of the performance capture volume respectively.  Function $\mathcal{F}$ is learnt using a CNN, specifically a convolutional Sec.~\ref{sec:autoenc} consisting of successive three-dimensional (3D) alternate convolutional filtering operations and down- or up-sampling with nonlinear activation layers for a similarly encoded output tensor $\mathbf{V_H}$, where $\mathbf{V_H} = \mathcal{F}(\mathbf{V_L}) = \mathcal{D}(\mathcal{E}(\mathbf{V_L}))$
%\begin{equation}
%\mathbf{V_H} = \mathcal{F}(\mathbf{V_L}) = \mathcal{D}(\mathcal{E}(\mathbf{V_L}))
%\end{equation}
for the learnt encoder ($\mathcal{E}$) and decoder ($\mathcal{D}$) functions. The encoder yields a latent feature representation via a series of 3D convolutions. Each convolutional layer is followed by batch normalisation and a ReLU in the Generator and convolutional strides for a layer in both the encoder and decoder. The encoder enforces $J(\mathbf{V_L}) = \mathcal{E}(\mathbf{V_L})$ where $J(\mathbf{V_L})$ is a concatenation of the skeletal pose vector corresponding to the input PVH; specifically a 78-D vector concatenation of 26 3D Cartesian joint coordinates in ${x, y,z}$ to generate the pose estimate and an additional latent embedding of size $\mathbf{e}$ (in general $\mathbf{e}=200)$. The decoder  inverts this process to output tensor $\mathbf{V_H}$ matching the input resolution but with higher fidelity. 
%\begin{figure}[htp]
%\centering
%\includegraphics[width=0.5\linewidth]{figures/ModelGenerator.jpg}
%   \caption{The 3D convolutional decoder-encoder Generator network with skip connections. %\label{fig:ModelGenerator}
%}
%\end{figure}
%Fig.~\ref{fig:ModelGenerator} illustrates the network architecture which incorporates two skip connections bypassing the network bottleneck to allow the output from a convolutional layer in the encoder to feed into the corresponding deconvolution layer in the decoder.  Combining the activations from the preceding layer in the main network and skip connection data via mean average, the use of mean average combination instead of element-wise addition or concatenation is analysed later in section~\ref{sec:skip}.
The full network parameters are: $n_{\mathcal{E}} =[64,64,128,128,256]$, $n_{\mathcal{D}} = [256,128,128,64,64]$, $k_{\mathcal{E}} = [3,3,3,3,3]$, $k_{\mathcal{D}}= [3,3,3,3,3]$, $k_s = [0,1,0,1,0]$ 
%\begin{table}[h!]
%\squeezeup
%%\centering
%%{
%%\small
%\begin{tabular}{l}
%%\hline
%$n_e$ =[64,64,128,128,256] \\
%$n_d$ = [256,128,128,64,64] \\
%$k_e$ = [3,3,3,3,3] \\
%$k_d$= [3,3,3,3,3] \\
%$k_s$ = [0,1,0,1,0] \\
%NumEpoch = 10 
%\end{tabular}
%}
%\label{tbl:netparams}
%\squeezeup
%\end{table}
where $k[i]$ indicates the kernel size and $n[i]$ is the number of filters at layer $i$ for the encoder ($\mathcal{E}$) and decoder ($\mathcal{D}$) parameters respectively. The location of the two skip connections are indicated by $s$ and link two groups of convolutional layers to their corresponding mirrored up-convolutional layer. The passed convolutional feature maps are averaged to the up-convolutional feature maps element-wise and passed to the next layer after rectification.  
%The central fully-connected layer encodes the $(78 + \mathbf{e})D$ latent representation ---discussed up in ln 212. 

The goal of $\mathcal{F}$ is thus to regress a high fidelity 3D volumetric representation given an initial poor fidelity blocky  3D volume estimate. Learning the end-to-end mapping from blocky volumes generated from a small number of camera viewpoints to both cleaner high fidelity volumes as if made by a greater number of camera viewpoints and accurate 3D joint position estimates, requires estimation of the weights $\phi$ in $\mathcal{F}$ represented by the convolutional and deconvolutional kernels. Specifically, given a collection of training sample triplets ${x^i, z^i, j^i}$, where $x^i \in \mathbf{V_L}$ is an instance of a low camera count volume, $z^i \in \mathbf{V_H}$ is the high camera count output groundtruth volume and $j^i \in \mathbf{J}$ is a vector of groundtruth joint positions for the given volume. The Mean Squared Error (MSE) is minimised at the output of the bottleneck and decoder across $N=W \times H \times D$ voxels through the two losses $\mathcal{L}_{joint}$ and $\mathcal{L}_{PVH}$.
%\begin{equation}
%\begin{split}
%\mathcal{L(\phi)} = &\mathcal{L}_{joint} + \lambda \mathcal{L}_{PVH}\\
%\mathcal{L(\phi)} = &\frac{1}{N}\sum^N_{i=1} \| \mathcal{F}(x^i: \phi) -z^i \|^2_2 +\lambda \mathcal{E}({\mathbf{V_L}}: \phi) -j^i \|^2_2 \\
%\end{split} \label{eq:DualLoss}
%\end{equation}

\begin{equation}
\mathcal{L(\phi)} = \mathcal{L}_{joint} + \lambda \mathcal{L}_{PVH} = \frac{1}{N}\sum^N_{i=1} \| \mathcal{F}(x^i: \phi) -z^i \|^2_2 +\lambda \mathcal{E}({\mathbf{V_L}}: \phi) -j^i \|^2_2 
\label{eq:DualLoss}
\end{equation}

%to add somewhere
%Human pose (joint positions) corresponding to the multi-view video frame is acquired using a commercial (Vicon Blade) human performance capture system run in parallel with video acquisition ( the TotalCapture and Human3.6M datasets provide such annotations).
Where $\lambda = 10^-3$, ensures both terms are of a similar magnitude.
%This scaling factor is analysed later in section~\ref{sec:ScalingFactor} exploring the use of different scaling factors.
\squeezeup
\squeezeup

\subsubsection{Generative Adversarial Network Model}
\label{sec:GAN}
The encoder-decoder model described in the section above with the dual volume and joint pose loss can produce consistent results. However, we propose to constrain and improve the reconstruction quality of the decoder output of the 3D occupancy volume and the pose estimation by employing a generative adversarial network (GAN).% Enforcing the learning of a realistic 3D occupancy volume with a discriminator loss, using the theory introduced by Goodfellow~\cite{goodfellow2014generative} defined as a \emph{game} between two competing networks: the Discriminator and the Generator. 

%The goal of the generative adversarial network is to recover a sharp, high-quality PVH volume $\mathbf{V_H}$, given a poor low-quality volume $\mathbf{V_L}$  with possible phantom parts, while simultaneously producing the 3D joint positions $J(\mathbf{V_L})$. 
The encoder model from section~\ref{sec:autoenc}, which we refer to as the \emph{Generator} $G$ estimates the improved volume, whilst the discriminator  maximises the chance of recognising real PVH volumes as real and generated PVH volumes as fake, optimizing the minimax objective::
\begin{equation}
\min_G \max_D V(D,G) = \mathbb{E}_{x \sim \mathbb{P}_r}[\text{log}(D(x))] +  \mathbb{E}_{\widetilde{x} \sim \mathbb{P}_g}[\text{log}(1 -D(\widetilde{x}))]
\end{equation}
where $P_r$ is the (real) data distribution and $P_g$ is the (generated) model distribution, defined by $\widetilde{x} = G(z), z \sim P(z)$, where the input $z$ is a sample from a simple noise distribution. Once both objective functions are defined, they are learnt jointly by the alternating gradient descent. 

%Initially, the decoder part of the Generator is pre-trained to learn a 3D pose estimate without the constraint of the 3D proxy, to produce the initial latent embedding. Once converged; alternatively, we train the Generator model’s parameters for a single iteration and fix the Discriminator’s parameters. Both networks are trained in alternating steps until the Generator produces good quality volume reconstructions using the dual loss found in Eq. ~\ref{eq:DualLoss}. An equal alternate training period for the network parts produced the most stable training process. 
%The Discriminator network 
%is shown in Fig.~\ref{fig:overview}, it 
%consists of 3 3D convolutional layers, which are followed by each time by batch normalisation and leaky ReLu activations.
%\begin{figure}[htp]
%\centering
%\includegraphics[width=0.3\linewidth]{figures/ModelDiscrimiantor.jpg}
%   \caption{The critic Discriminator network on the volume reconstruction\label{fig:ModelDiscrimiantor}.
%}
%\end{figure}
\squeezeup
\squeezeup

\subsubsection{Skip Connections}
Deeper networks in image restoration tasks can result in finer image details being lost given the compact latent space. Recovery of this detail is an under-determined problem, exasperated by the need to reconstruct the additional dimension in volumetric data. We add skip connections between two corresponding convolutional and deconvolutional layers. 
%Allowing intermediate stages of the encoder to transmit directly to latter stages of the decoder can aid the reconstruction of high-frequency detail and mitigate the vanishing gradient problem of many-layered networks by providing a new direct route for the error gradient to back-propagate to early layers. 
%Our proposed skip connections differ from that proposed in recent image restoration work \cite{srivastava2015trainingSkip,he2016ResNet} which concern only smoother optimisation. Instead, we pass the feature activation’s at intervals of every two convolutional layers to their mirrored up-convolutional layers to enhance reconstruction detail. 
Omitting the skip connections the detail of extremities such as lower arm position is poorly estimated by both the volume and 3d joints (see sup. material).
\squeezeup
\squeezeup

\subsubsection{Temporal Consistency}
\label{sec:TempConsistency}
Given the inherent temporal nature of the human pose, we enforce temporal consistency with additional Long Short Term Memory (LSTM) layers. These help to smooth noisy individual joint detections to enable a smoother prediction of the joint estimation. The latent vector from the encoder $J(\mathbf{V_L}_t) = \mathcal{E}(\mathbf{V_L}_t)$ at time $t$ consisting of concatenated joint spatial coordinates passed through a series of gates resulting in an output joint vector $\mathbb{J}_o$. The aim is to learn the function that minimises the loss between the input vector and the output vector $\mathbb{J}_o = o_t \circ tanh(c_t)$ ($\circ$ denotes the Hadamard product) where $o_t$ is the output gate, and $c_t$ is the memory cell, a combination of the previous memory $c_{t-1}$ multiplied by a decay based forget gate, and the input gate. Thus, intuitively the LSTM result is the combination of the previous memory and the new input vector. In this implementation, the model consists of two LSTM layers both with 1024 memory cells, using a look back of $T = 5$.

\squeezeup
\squeezeup
\section{Results and Discussion}

%\subsection{Experimental Setup}
\label{sec:TCSetup}
To quantify the performance of our proposed approach, we report \emph{Mean Per Joint Position Error}, the mean 3D Euclidean distance between ground-truth and estimated joint positions of the 26 joints. We performed quantitative evaluation over two public multi-view video datasets of human actions. 3D human pose is evaluated for Human 3.6M~\cite{h36m_pami}, and the performance of both the skeleton estimation and volume reconstruction is evaluated on TotalCapture~\cite{trumble_total_2017}.
%and TotalCaptureOutdoor~\cite{Malleson3DV17} datasets. 

%including hips, knees, ankles, neck, head, shoulders, elbows and wrists.  In order to evaluate pose accuracy independently of absolute camera position and orientation, we align our estimates with the ground-truth. Aligning with the ground truth is standard practice in existing benchmarks~\cite{h36m_pami,TrumbleBMVC17}. Thus, in our case, the Mean Per Joint Position Error is a measure of pose accuracy independent of global position and orientation.
%To generate both training and test sequences, given the temporal requirement of the LSTM, we translated a sliding window of length $T$ successively by a single frame across the sequence. Hence there is an overlap between the frames, providing additional data to train on, which is always an advantage for deep learning systems. While during test time, we initially predict the first $T$ frames of the sequence and slide the window by a stride length of 1 to predict the next frame using the previous pose $T$ estimates.
To train $\mathcal{F}$, 
 %we use Adadelta~\cite{zeiler2012adadelta} an extension of Adagrad that seeks to reduce it’s aggressive, radically diminishing learning rates, restricting the window of accumulated past gradients to some fixed size $w$. 
 we initially, train the encoder for just the skeleton loss, purely as a pose regression task without the decoder or critic networks, due to the large parameter count in the volumetric network. These trained weights initialise the encoder stage to help constrain the latent representation during the full, dual-loss network training. Then given the learnt weights as initialisation for the encoder section, we train the entire encoder/decoder network end-to-end constrained by the dual loss of the skeleton and volume occupancy through the GAN critic network. The encoder-decoder Generator and Discriminator network are trained alternately, with the opposing network weights fixed. 
 %with training typically converging within ??  iterations, 
 
 %A factor of $\lambda$ scales the pose term of the dual loss (Eq.~\ref{eq:DualLoss}). We found the approach insensitive to this parameter up to an order of magnitude and set $\lambda=10^{-3}$ for all experiments. Below $10^{-3}$, the bottleneck convergences to a semantic representation of the pose that is stable but does not resemble joint angles while above $10^{-2}$ the network will not converge. 
 %For training the network, a learning rate of ?? is used that is decayed exponentially per iteration. 
We train with a batch size of 32 and a sequence length of $T=5$ (we experimented with different sequence lengths and found sequence length 3, 4, 5 and 6 generally gave similar results). We augment the data during training with a random rotation around the central vertical axis of the PVH to introduce rotation invariance. 
 %A single training step for sequences of length 5 takes only ?? ms approximately, while a forward pass takes only about ??ms on NVIDIA Titan XP GPU. Therefore, given the 2D joint locations from a pose detector, our network takes about ??ms to predict 3D pose per frame.
\squeezeup
\squeezeup

\subsection{TotalCapture Evaluation}
\begin{table*}[htb]
\centering
{
\small
\begin{tabular}{lcccccccc}
\hline
%% after \\: \hline or \cline{col1-col2} \cline{col3-col4} ...
Approach &                              Num    &\multicolumn{3}{c}{SeenSubjects(S1,2,3)}&\multicolumn{3}{c}{UnseenSubjects(S4,5)} & Mean \\
                                         &Cams& W2 & FS3 & A3 & W2 & FS3 & A3 & \\ \hline
%Tri-CPM~\cite{cao2016realtimeCPM}           & 79.0 & 112.1 &106.5& 79.0 &149.3 & 73.7 & 99.8 \\
Tri-CPM-LSTM~\cite{cao2016realtimeCPM}   & 8  & 45.7 &102.8 & 71.9& 57.8 & 142.9 & 59.6 & 80.1 \\ 
%2D Matte~\cite{TrumbleCVMP2DConvNet}    & 104.9 & 155.0 & 117.8 & 161.3 & 208.2 & 161.3 & 142.9 \\
2D Matte-LSTM~\cite{TrumbleCVMP2DConvNet}& 8  & 94.1 &128.9  &105.3 & 109.1& 168.5&120.6&121.1 \\ 
%3D PVH                                  & 48.3 & 122.3  & 94.3& 84.3 &168.5 & 154.5& 107.3\\
%3D PVH-TSP                              & 38.8 &{\bf 86.3}  & 72.6    & 69.1 & 112.9& 119.5     & 81.1\\
3D-PVH~\cite{trumble_total_2017}         & 8+13 IMU& 30.0 & 90.6 & 49.0 & 36.0 & 112.1 & 109.2 & 70.0 \\ 
AutoEnc~\cite{trumble:eccv:2018}         & 8 & 13.4 & 49.8 & 24.3 & 22.0 & 71.7 & 40.7 & 35.5 \\ 
Fusion-RPSM~\cite{Qiu:iccv:2019}         & 8 & 19   &58    &21    &32    &54    &33   & 29 \\
IMU 1Cam SMPL~\cite{PonsECCV18}          &1+13 IMU& -    &   -  &  -   &      &  -   &   -  & 26.0 \\ \hline
Proposed DualLoss GAN                    & 2 & 9.2 & 30.3 & 15.2 & 13.3 & 41.7 & 25.3 & 21.4 \\ \hline
\end{tabular}
}
\caption{Comparison of our approach on  TotalCapture  to other human pose estimation approaches, expressed as average per joint error (mm) on previously seen and unseen test subjects. (where W2, FS3, A3 are groups of test sequences of walking, freestyle and acting respectively)}
\label{tab:totalcaptureResults}
\squeezeup
\end{table*}
\label{sec:TCEval}
%\begin{figure}[htb]
%\centering
%\includegraphics[width=1\textwidth]{figures/TC360Views.jpg}
%\caption{Reconstruction of a challenging test pose from S5 Freestyle3 (Frame 2679) rendered from 10 viewpoints.}
%\label{fig:TC360Cam}
%\end{figure}

%\begin{figure}[htb]
%\centering
%\includegraphics[width=1\linewidth]{figures/FrameWiseTC.jpg}
%\caption{The frame-wise mean joint error for all sequences in the TotalCapture dataset, (error in y-axis in mm, frame on the x-axis).}
%\label{fig:FrameWiseTC}
%\end{figure}
We quantitatively evaluate tracking accuracy on the TotalCapture dataset~\cite{trumble_total_2017}. 
%The dataset consists of 5 subjects performing several activities such as walking, acting, a range of motion sequence (ROM) and freestyle motions, which are recorded using 8 calibrated static HD RGB cameras. 
%The dataset has publicly released foreground mattes that we use to compute the occupancy PVH, and we use the released RGB images to localise the semantic 2D joint estimates. We obtain ground-truth poses using a marker-based motion capture system, with the markers are $<5$mm in size and therefore invisible to the training model. 
%All data is synchronised and operates at a framerate of 60Hz, providing ground truth poses as joint positions. 
We study the accuracy gain due to our method by ablating the set of camera views available on the {\em TotalCapture} dataset.  Jointly training the generative adversarial dual loss model using high fidelity PVHs obtained using all ($C=8$) views of the dataset and 78-D vector concatenation of the 26 3D Cartesian pose joint coordinates. With the corresponding input low fidelity, PVHs obtained using fewer views (we train for $C=2$ and $C=4$ random neighbouring views), we follow the train and test strategy of~\cite{trumble_total_2017}. The dataset contains five subjects, with four diverse categories of sequences; \emph{ROM, Walking, Acting, and Freestyle}, with each sequence, repeated three times by each subject. The sequences are long, with around 3000-5000 frames, resulting in 1.9M frames. Within the acting and freestyle sequences, there is a great deal of diversity in the captured content.
%The model is then tested on held-out footage to determine the degree to which it can estimate the 3D pose and reconstruct a high fidelity PVH from the ablated set of camera views. The dataset consists of a total of four male and one female subjects each performing four diverse performances, repeated three times: \emph{ROM, Walking, Acting and Freestyle}, and each sequence lasts around 3000-5000 frames. Forming the train and test partitions with respect to the subjects and sequences, the training consists of {\em ROM1,2,3}; {\em Walking1,3}; {\em Freestyle1,2} and {\em Acting1,2} on subjects 1,2 and 3. The test set is the performances Freestyle3 (\textbf{FS3}), Acting (\textbf{A3}) and Walking2 (\textbf{W2}) on subjects 1,2,3,4 and 5. This split 
%to allows for separate evaluation on unseen and on seen subjects but always on unseen sequences. 

 The PVH at $C=8$ provides the ideal 3D reconstruction proxy estimation for comparison, while $C=\{2,4\}$ input covers at most a narrow $90^\circ$ view of the scene.  Before refinement, the ablated view PVH data exhibits phantom extremities and lacks fine-grained detail, particularly at $C=2$ (Fig.~\ref{fig:TCQuail2CamResults}). These crude volumes would be unsuitable for pose estimation or reconstruction as they do not reflect the true geometry and would cause poor defined joint estimations and severe visual misalignments when projecting camera texture onto the model. However, our method can estimate the joint positions accurately and also clean up and hallucinate a volume equivalent to one produced by the unabated $C=8$ camera viewpoints. Tab.~\ref{tab:totalcaptureResults} quantifies the pose animation error between previous approaches using in general multiple camera views~\cite{cao2016realtimeCPM,TrumbleCVMP2DConvNet,trumble_total_2017,trumble:eccv:2018} or additional data modalities~\cite{trumble_total_2017,PonsECCV18} and our proposed approach with only two camera views. 
 %We outperform the single loss learning-based approach introduced in the TotalCapture dataset~\cite{trumble_total_2017} by 48mm, this approach uses all eight cameras and fuses the data of 13 IMU sensors with the probabilistic visual hull. 
 %The approach of Pons~\cite{PonsECCV18} also uses the 13 IMUs sensors and a single reference camera and achieve similar performance to us of 26mm; however, it requires that the full sequence is simultaneously optimised over. We do not require sensors to be placed on the human, removing the requirement to pre-makeup the subject and only require an additional camera and receive a similar low joint error. 
 We outperform best camera approach~\cite{Qiu:iccv:2019} by 8 mm indicating the importance of the GAN loss and semantic 2D joint estimates.
 
% Figs.~\ref{fig:TCResults},\ref{fig:TC360Cam} illustrates the performance of the approach qualitatively on challenging frames (see also the supplementary video ).  
 %Fig.~\ref{fig:FrameWiseTC} illustrates the frame-wise error across all frames of all sequences in the TotalCapture dataset, and our approach can maintain a low, with no dramatic failure frames present, with the maximum mean error of only ~7cm and mean of only 2cm. 
 %Error peaks are generally caused by a simultaneous failure of both channels of the PVH, the foreground occupancy and 2D semantic joints. For example, missing or weakly defined limb extremities, and such data is under-represented within the training data, the error is otherwise consistently low. 
\begin{figure}[htb]
\centering
\includegraphics[width=0.9\linewidth]{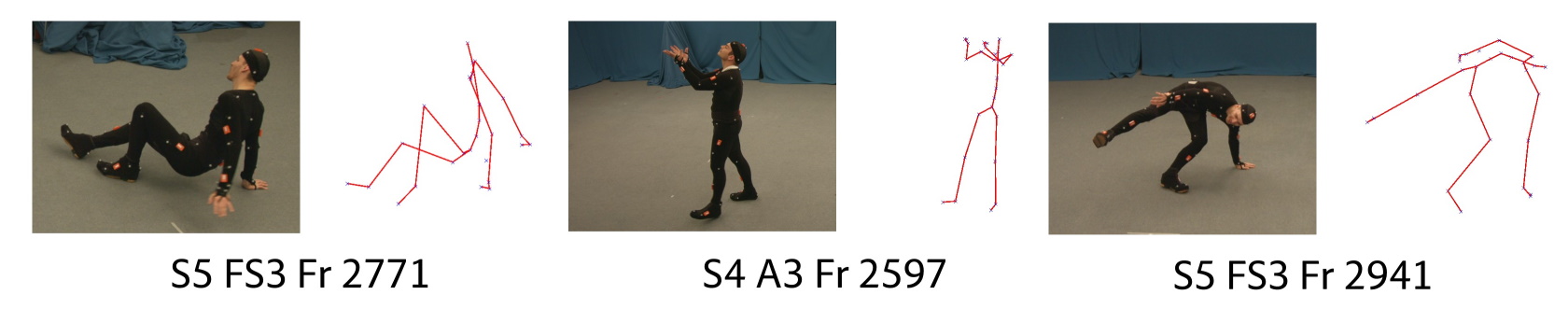}
\caption{Representative pose estimations from (Fr)ames of unseen (S)ubjects performing (A)ctions with challenging poses (TotalCapture dataset).}
\label{fig:TCResults}
\end{figure}

\squeezeup
\squeezeup
\subsection{Ablation Study}
\begin{figure}[t!]
\centering
\includegraphics[width=0.9\linewidth,height=4.5cm]{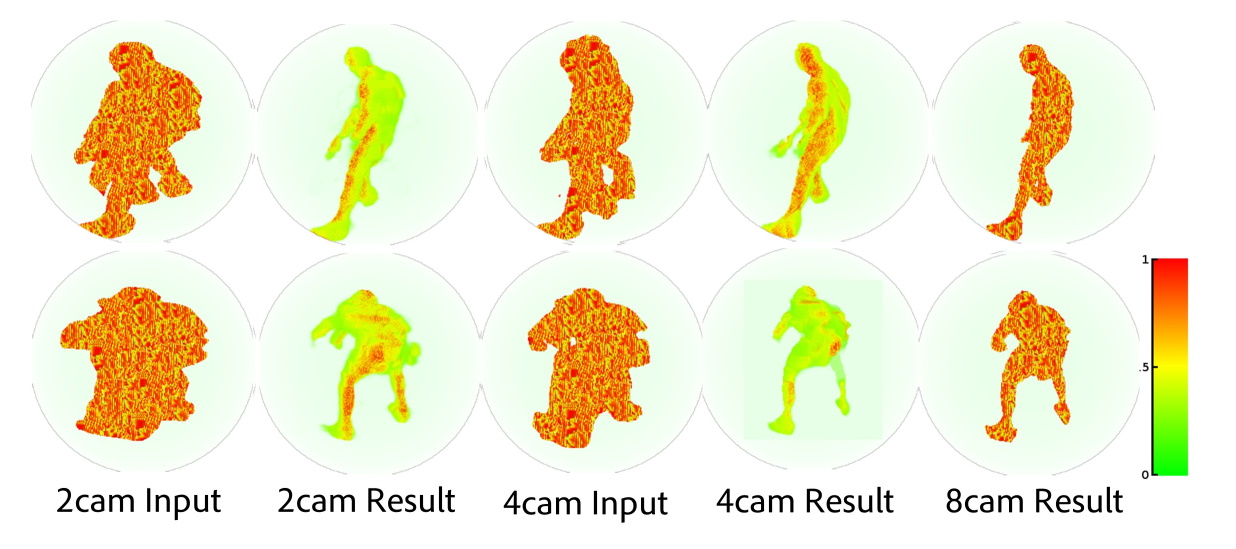}
\caption{Examples of input/resultant reconstructions for [2,4,8] cameras on TotalCapture.}
\label{fig:NuMCamsPVH}
\squeezeup
\squeezeup
\end{figure}

To understand the influence of the individual components and design decisions, we perform an ablative analysis of tracking accuracy for our individual contributions (Tab.~\ref{tab:totalcaptureSplitResults}).
\begin{table*}[t]
\begin{adjustbox}{width=1.0\textwidth}
\centering

\small
\begin{tabular}{lccccccccccccc}

\hline
%% after \\: \hline or \cline{col1-col2} \cline{col3-col4} ...
Approach &\multicolumn{2}{c}{Features}&\multicolumn{4}{c}{Model}& \multicolumn{3}{c}{SeenSubjects(S1,2,3)}&\multicolumn{3}{c}{UnseenSubjects(S4,5)} & Mean \\
                    &Occ  &2Djoint&Enc    &Dec    &LSTM&GAN   &W2    &FS3   & A3   &W2    &FS3    & A3 & \\ \hline
Encoder             &8cam &   -   &\tick  & -     & -  & -    & 42.0 & 120.5& 59.8 & 58.4 & 162.1 & 103.4 & 85.4  \\
EncoderLSTM         &8cam &  -    &\tick  &-      &\tick& -   & 15.2 &65.7  &54.4  &17.8  & 73.0 & 50.6 &  58.4   \\
AutoEncLSTM         &8cam & -     &\tick  &\tick  &\tick& -   & 13.4 & 49.8 & 24.3 & 22.0 & 71.7 & 40.7 & 35.5    \\
2DJoint             &-    & 8cam  &\tick  &\tick  &\tick& -   & 21.2 & 123.1& 88.6 & 105.7&142.2 & 97.7 & 41.2    \\
Occ+2DJoint         &8cam & 8cam  &\tick  &\tick  &\tick& -   & 10.2 & 123.1& 88.6 & 105.7&142.2 & 97.7 & 31.1    \\ \hline
GAN8cam            &8cam &8cam   &\tick  &\tick  &\tick&\tick& 8.2 & 30.5 & 15.0  & 10.2 & 40.8 & 24.7 & 20.7 \\ 
GAN4cam            &4cam &4cam   &\tick  &\tick  &\tick&\tick& 9.8 & 29.9 & 15.3  & 13.5 & 42.2 & 24.9 & 21.6 \\ 
GAN2cam            &2cam &2cam   &\tick  &\tick  &\tick&\tick& 9.2 & 30.3 & 15.2  & 13.3 & 41.7 & 25.3 & 21.4 \\ \hline

\end{tabular}
\end{adjustbox}

\caption{Ablation study of the Mean per joint error (mm). for the individual components on the TotalCapture dataset.}
\label{tab:totalcaptureSplitResults}
\squeezeup
\end{table*}
Each part of the process enables an improvement in the accuracy performance, especially the use of temporal information (\textbf{EncoderLSTM)} and dual loss in the approach (\textbf{AutoEncLSTM}). The inclusion of the 2D joint (\textbf{2DJoint}) estimates into the dual-channel PVH further reduces this loss by around 4 mm to 31.1 average joint error. The inclusion of the Discriminator (\textbf{GAN8cam}) to enforce improved 3D occupancy volume result, enables the loss to be further reduced to 21mm per joint using all eight camera views. The greater the number of cameras, the more visually realistic the input PVH is. However, it is possible to remove a large number of these cameras with little or no impact on performance (\textbf{GAN4cam} and \textbf{GAN2cam}). Despite greatly degrading the appearance of the input PVH when using only 2 or 4 views as input, as indicated by Fig.~\ref{fig:NuMCamsPVH}. The figure also illustrates the resulting output PVH, and this can be seen to be of a high-fidelity result invariant to the number of cameras used.
In summary, using a low fidelity PVH  from only two camera views with phantom and missing voxels, achieves a headline performance of 21.4mm mean per joint error.

\subsection{Evaluating Reconstruction Accuracy}

In addition to the pose estimation, the dual loss model is also able to reconstruct the high-fidelity 3D volume for the given low fidelity PVH input. Tab.~\ref{tab:QuantTC} quantifies the error between the unablated ($C=8$) and the reconstructed volumes for $C=\{2,4\}$ view PVH data, baselining these against $C=\{2,4\}$ PVH prior to enhancement via our learnt model (\emph{input}). 

\begin{figure}[htb]
\centering
\includegraphics[width=0.8\linewidth]{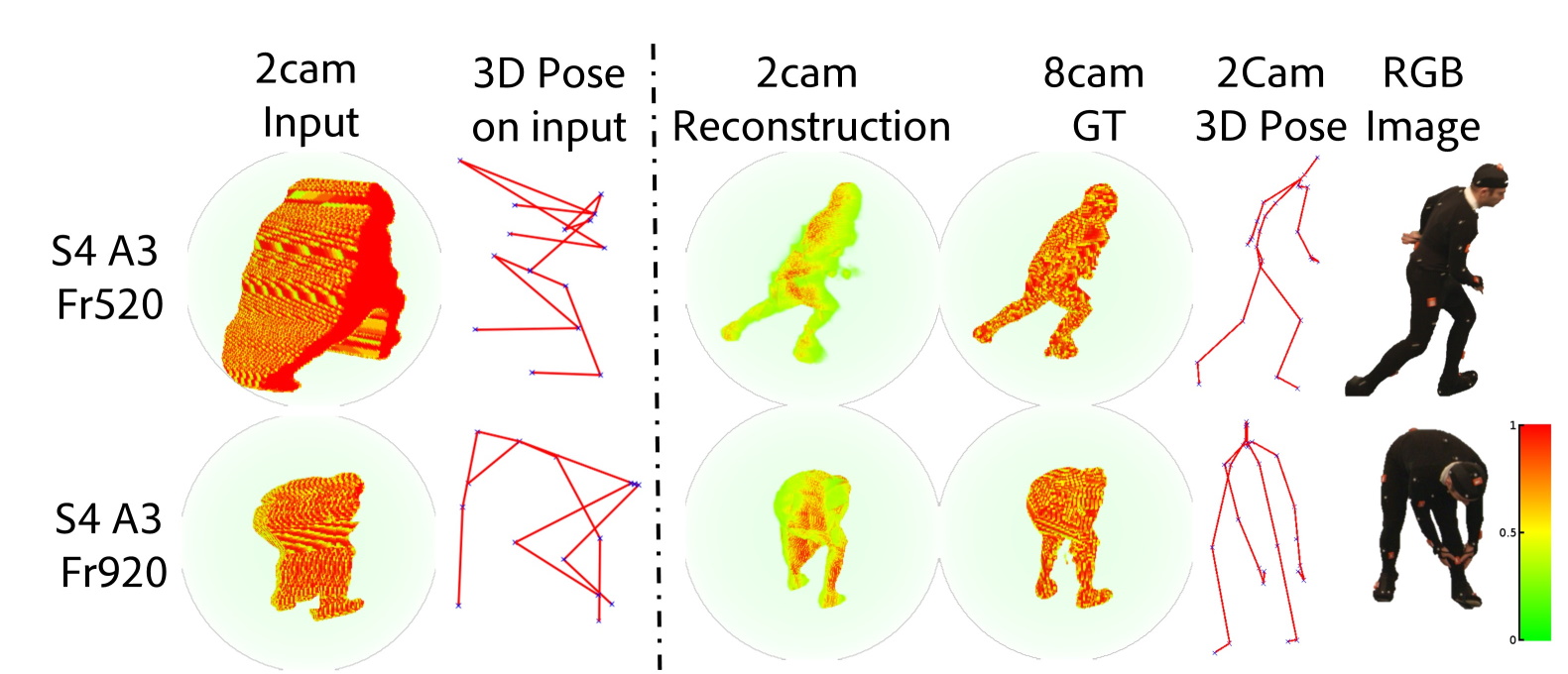}
\caption{Qualitative visual comparison of the input PVH and 3D Pose estimate on encoder against the resultant Reconstruction and 3D Pose estimation using $C=\{2\}$ views on the TotalCapture dataset. False colour volume occupancy (PVH) and groundtruth $C=8$ PVH.}
\label{fig:TCQuail2CamResults}
\end{figure}

\begin{table}[htb]
\centering
{
\small
\begin{tabular}{lcccccccc}
\hline
%% after \\: \hline or \cline{col1-col2} \cline{col3-col4} ...
Method&Cams &\multicolumn{3}{c}{SeenSubs(S1,2,3)}&\multicolumn{3}{c}{UnseenSubs(S4,5)} & Mean   \\
                             & C&W2  &FS3  & A3  &W2  &FS3  &A3  &       \\ \hline
Input                    &2 &19.1&28.5 &23.9 &23.4&27.5 &25.2&24.6   \\
Input                     & 4&11.4&16.5 &12.5 &12.0& 15.2&14.2&11.6  \\ \hline
~\cite{gilbert2018volumetric}& 2&5.43&10.03&6.70 &5.34&10.05&8.71&7.71 \\ \hline
Ours                     &2 &5.44& 9.94&6.34 &5.16&9.86 &8.49&7.34  \\   
Ours                     &4 &4.85& 9.32&5.84 &4.83&9.56 &8.03&7.02  \\   \hline
\end{tabular}
}
\caption{ Quantitative performance of volumetric reconstruction on the TotalCapture dataset using 2-4 cameras before our approach (Input) and after, versus unablated groundtruth using eight cameras (error as MSE $\times 
10^{-3}$).  Our method reduces reconstruction error to 30\% of the baseline (Input) for two views.}
%\squeezeup
\label{tab:QuantTC}
\end{table}
To measure the performance, we compute the average per-frame MSE of the probability of occupancy across each sequence. Comparing the two and four camera PVH volume before enhancement and our results indicate a reduction in MSE of around three times through our approach when using two cameras views for the input and a halving of MSE for a PVH formed from 4 cameras. View count $C=4$ in a $180^\circ$ arc around the subject perform slightly better than $C=2$ neighbouring views in a $90^\circ$ arc. However, the performance decrease is minimal for the significantly increased operational flexibility that a two camera deployment provides. In all cases, MSE is more than halved (up to 34\% lower) using our refined PVH for a reduced number of views.  Using only two cameras, we can produce an equal volume to that reconstructed from a full $360^\circ$ $C=8$ setup. We show qualitative results of using only two and four camera viewpoint to construct the volume in Fig.~\ref{fig:TCQuail2CamResults}.

\squeezeup
\squeezeup

\subsection{Human 3.6M evaluation}
We perform a further quantitative and qualitative evaluation on the Human 3.6M~\cite{h36m_pami} dataset. Human 3.6M is the largest publicly available dataset for human 3D pose estimation and contains 3.6 million images of 7 different professional actors performing 15 everyday activities. Each video is captured using four calibrated cameras arranged in the $360^\circ$ arrangement and contains 3D pose ground truth. We follow the standard train and evaluation protocols of the Human3.6M dataset ~\cite{li2015maximumH36m,tome2017liftingH36m}.
%, using subjects 1, 5, 6, 7, and 8 for training, and subjects 9 and 11 for testing. The error is evaluated like that reported for the TotalCapture dataset on the predicted 3D pose without any transformation. 
 Therefore, we explore (Tab.~\ref{tab:H36mResults})
the transfer of the high fidelity 8cam trained model from the TotalCapture dataset to the 4 cam human3.6M dataset through three specified methods of training:
\begin{table*}[htb]
%\begin{adjustbox}{width=1.0\textwidth}
\centering
{
\small
\begin{tabular}{lcccccccc}
\hline
Approach                              & Direct. & Discus & Eat & Greet. &Phone &Photo &Pose &Purch. \\ \hline
Lin \emph{et al}~\cite{li2015maximumH36m}          & 132.7& 183.6 & 132.4& 164.4 &162.1&205.9&150.6 &171.3 \\
%Tekin~\cite{tekinapredictingH36m}     & 102.39 & 136.88 & 87.95 & 126.83 &185.02&185.02&114.69 &107.61 \\
%ekin~\cite{tekin2016fusingH36m}       & 85.0 & 108.8 & 84.4 & 98.9 &119.4&95.7 &98.5 &93.8 \\
%Zhou~\cite{zhou2016sparsenessH36m}    & 87.36 & 109.31 & 87.05 & 103.16 &116.18&143.32&106.88 &99.78 \\
%Sanzari~\cite{sanzari2016bayesianH36m}&48.82& 56.31 & 95.98 & 84.78 & 96.47&105.58&66.30 &107.41 \\
%Tome~\cite{tome2017liftingH36m}       &65.0 & 73.5 & 76.8 & 86.4 & 86.3&110.7&68.9 &74.8 \\ 
%Trumble~\cite{trumble_total_2017}     & 92.7& 85.9& 72.3& 93.2& 86.2&101.2 &75.1 &78.0 \\ 
Lin \emph{et al}~\cite{lin2017CVPRRPSM}              &58.0    & 68.3 & 63.3 &65.8  & 75.3 &93.1&61.2&65.7 \\ 
%Martinez~\cite{martinez_simple_2017}          &51.8 & 56.2&58.1 & 59.0& 69.5&78.4  & 55.2 & 58.1 \\
%Tri-CPM~\cite{cao2016realtimeCPM}     & 125.0 & 111.4 & 101.9 & 142.2 &125.4 & 147.6& 109.1 & 133.1 \\
%Tri-CPM-TSP~\cite{cao2016realtimeCPM}     &67.4& 71.9 & 65.1 & 108.8 & 88.9 & 112.0&55.6 & 77.5 \\ \hline
%PVH                                   & 152.8& 171.4& 152.6& 189.2& 179.7& 210.2& 147.1& 167.0 \\
Trumble \emph{et al}~\cite{trumble:eccv:2018}        & 41.7& 43.2& 52.9& 70.0& 64.9& 83.0& 57.3& 63.5\\ 
Imtiaz \emph{et al}~\cite{hossain2018exploiting}     &44.2 &46.7 &52.3 &49.3 &59.9 &59.4 &47.5 &46.2 \\  
Qiu \emph{et al}~\cite{Qiu:iccv:2019}                &28.9 &32.5 &26.6& 28.1 &28.3 &29.3 & 28.0& 36.8 \\ \hline
Human3.6Model                           &55.6 &52.1 &51.8 &59.9 &62.1 &58.2 &55.2 &62.0                      \\
TCModel                       &37.1 &45.3 &47.1 &45.9 &60.1 &57.6 &49.9 &48.1                       \\
TCModel+FineTune(H36M)             &26.0 &24.0 &23.5 & 23.5 & 33.3 & 38.2 & 27.1 &25.2 \\ \hline

                                        & Sit. & Sit D & Smke & Wait &W.Dog& walk & W. toget. &Mean\\\hline
Lin \emph{et al}~\cite{li2015maximumH36m}            & 151.6 & 243.0 & 162.1 &170.7 &177.1& 96.6 & 127.9 & 162.1 \\
%Tekin~\cite{tekinapredictingH36m} & 136.15 &205.65 & 118.21 &146.66 &128.11& 65.86 & 77.21 & 125.28 \\
%ekin~\cite{tekin2016fusingH36m}         & 73.8 & 170.4 &85.1 &116.9 &113.7& 62.1 & 94.8 & 100.1 \\
%Zhou~\cite{zhou2016sparsenessH36m}& 124.52 & 199.23 & 107.42 &118.09 &114.23& 79.39 &97.70 \& 113.01 \\
%Sanzari~\cite{sanzari2016bayesianH36m}&116.89&129.63 &97.84 &65.94 &130.46&92.58 &102.21 &93.15 \\
%Tome~\cite{tome2017liftingH36m}         & 110.2 &173.9 &85.0 &85.8 &86.3 &71.4 &73.1 & 88.4 \\ 
%Tri-CPM~\cite{cao2016realtimeCPM}     & 135.7 & 142.1 & 116.8 & 128.9 &111.2 & 105.2 & 124.2 & 124.0 \\
%Tri-CPM-TSP~\cite{cao2016realtimeCPM}     & 92.7 & 110.2 & 80.3 & 100.6 &71.7 & 57.2 & 77.6 &88.1 \\ 
%Trumble~\cite{trumble_total_2017}       & 83.5& 94.8& 85.8&82.0 &114.6 &94.9 &79.7 &87.3 \\
Lin \emph{et al}~\cite{lin2017CVPRRPSM}              &98.7 &127.7 &70.4 &68.2   & 73.0 & 50.6 & 57.7 & 73.1 \\ 
%Martinez~\cite{martinez_simple_2017}    & 74.0 &94.6&62.3&59.1&65.1   &49.5 &52.4 &62.9 \\
%PVH \& 177.3& 192.8& 179.3& 161.0& 236.8& 179.0& 168.8& 169.0 \\
Trumble \emph{et al}~\cite{trumble:eccv:2018}        &61.0 &95.0 &70.0 &62.3 &66.2 &53.7 &52.4 &62.5 \\     
Imtiaz \emph{et al}~\cite{hossain2018exploiting}     &59.9 & 65.6 &55.8 &50.4 &52.3 &43.5 &45.1 &51.9 \\     
Qiu \emph{et al}~\cite{Qiu:iccv:2019}                &42.0 & 30.5 &35.6 &30.0 &28.3 &30.0 &30.5 &31.2 \\ \hline
Human3.6Model                           &53.3 & 74.6 &61.8 & 59.1 &61.8 & 65.8 & 61.2 & 59.6  \\   
TCModel                       &56.8 &68.2 & 56.3 & 53.1 &47.7 & 50.5 & 50.2 & 54.7   \\
TCModel+FineTune(H36M)             &30.2 &48.1 & 37.6 &31.2 & 34.4 & 28.1 & 27.1 & 30.5     \\ \hline
\end{tabular}}
%\end{adjustbox}
\caption{Comparison of the proposed 3 methods to baseline methods on  Human 3.6M.}
\label{tab:H36mResults}
\squeezeup
\end{table*}

\noindent {\textbf{Human3.6Model:}  A baseline approach, using the specified Human 3.6M training data with the four cameras assuming the semantic 2D joints will compensate in part for the phantom part and ghosting that occurs to the occupancy voxels.} \\
\noindent {\textbf{TCModel:} Transfer of the trained $2 \mapsto 8$ camera views model from the TotalCapture dataset, without any further training, to estimate pose as if 8 cameras were used at acquisition.}\\
\noindent {\textbf{TCModel+FineTune(H36M):}  2 epochs of fine-tuning of the learnt $2 \mapsto 8$  TCModel on Human3.6M dataset.}\\
%together with further recent approaches and qualitative example frames shown are shown in Fig.~\ref{fig:H36mEx}.
Our TotalCapture trained model (\textbf{TotalCaptureModel)} improves the baseline training of Human 3.6M (\textbf{Human3.6Model}) alone by 5mm and the combined TotalCapture of fine-tuned model \textbf{TotalCapture+FineTune(H36M Model)} improves this performance by a further 10mm. Our network improves on  Qiu~\cite{Qiu:iccv:2019}, and dramatically improves on other prior work. By using the information of temporal context and semantic joint estimations, our network reduces the overall error in estimating 3D joint locations, especially on actions like phone, photo, sit and sitting down on which for previous methods did not perform well due to heavy occlusion. 
\squeezeup
\squeezeup
%%%%%%%%%%%%%%%%%%%%%%%%%%%%%%%%%%%%%%%%%%%%%
%\subsection{Supplementary analysis on TotalCapture, Human 3.6M and TotalCaptureOutdoor}
%\label{sec:evalExtra}
%In the supplementary material, we explore the effect of increasing the angle between the cameras to vary the camera layout, the input PVH resolution, training data quantity and quality and the effects of the skip connections. Together with qualitative tests on the outdoor dataset TotalCaptureOutdoor~\cite{Malleson3DV17}.

%\squeezeup
%\squeezeup

\section{Conclusions}
This proposed work generates accurate 3D joint and 3D volume proxy reconstructions, from a minimal set of only two wide baseline cameras, through learning constrained by a dual loss on the joints and a generative adversarial loss on the 3D volume. The dual loss in conjunction with the Discriminator in the GAN framework delivers state of the art performance. Furthermore, we have demonstrated that a trained model with plentiful data (from the TotalCapture dataset) can be used to improve performance on other sets of data (in this case from the Human3.6M dataset) that have a limited set of camera views. 

%\section*{Acknowledgements}

%InnovateUK supported the work via the Total Capture project, grant agreement 102685 and through the donation of GPU hardware by Nvidia.

%BibTeX users please use one of
\bibliographystyle{spmpsci}      % mathematics and physical sciences
%\bibliography{Pose.bib}   % name your BibTeX data base

\begin{thebibliography}{47}
\providecommand{\natexlab}[1]{#1}
\providecommand{\url}[1]{\texttt{#1}}
\expandafter\ifx\csname urlstyle\endcsname\relax
  \providecommand{\doi}[1]{doi: #1}\else
  \providecommand{\doi}{doi: \begingroup \urlstyle{rm}\Url}\fi

\bibitem[Abrahamsson et~al.(2017)Abrahamsson, Blom, and Jans]{Abrahamsson2017}
S.~Abrahamsson, H.~Blom, and D.~Jans.
\newblock Multifocus structured illumination microscopy for fast volumetric
  super-resolution imaging.
\newblock \emph{Biomedical Optics Express}, 8\penalty0 (9):\penalty0
  4135--4140, 2017.

\bibitem[Andriluka et~al.(2009)Andriluka, Roth, and Schiele]{andriluka09}
M.~Andriluka, S.~Roth, and B.~Schiele.
\newblock Pictoral structures revisited: People detection and articulated pose
  estimation.
\newblock In \emph{Proc. Computer Vision and Pattern Recognition}, 2009.

\bibitem[Andriluka et~al.(2014)Andriluka, Pishchulin, Gehler, and
  Schiele]{andriluka2014MPI2DPoseDataset}
Mykhaylo Andriluka, Leonid Pishchulin, Peter Gehler, and Bernt Schiele.
\newblock 2d human pose estimation: New benchmark and state of the art
  analysis.
\newblock In \emph{Proceedings of the IEEE Conference on Computer Vision and
  Pattern Recognition}, pages 3686--3693, 2014.

\bibitem[Aydin and Foroosh(2017)]{Atalay2017}
V.~Aydin and H.~Foroosh.
\newblock Volumetric super-resolution of multispectral data.
\newblock In \emph{Corr. arXiv:1705.05745v1}, 2017.

\bibitem[Cao et~al.(2016)Cao, Simon, Wei, and Sheikh]{cao2016realtimeCPM}
Zhe Cao, Tomas Simon, Shih-En Wei, and Yaser Sheikh.
\newblock Realtime multi-person 2d pose estimation using part affinity fields.
\newblock \emph{ECCV'16}, 2016.

\bibitem[Cao et~al.(2017)Cao, Simon, Wei, and Sheikh]{cao2017realtime}
Zhe Cao, Tomas Simon, Shih-En Wei, and Yaser Sheikh.
\newblock Realtime multi-person 2d pose estimation using part affinity fields.
\newblock In \emph{CVPR}, 2017.

\bibitem[Casas et~al.(2015)Casas, Huang, and Hilton]{casas2014rwvc}
Dan Casas, Peng Huang, and Adrian Hilton.
\newblock {Surface-based Character Animation}.
\newblock In Marcus Magnor, Oliver Grau, Olga Sorkine-Hornung, and Christian
  Theobalt, editors, \emph{Digital Representations of the Real World: How to
  Capture, Model, and Render Visual Reality}, chapter~16, pages 239--252. {CRC}
  Press, April 2015.
\newblock ISBN 9781482243819.

\bibitem[Collet et~al.(2015)Collet, Chuang, Sweeney, Gillett, Evseev,
  Calabrese, Hoppe, Kirk, and Sullivan]{collet2015MSFVV}
Alvaro Collet, Ming Chuang, Pat Sweeney, Don Gillett, Dennis Evseev, David
  Calabrese, Hugues Hoppe, Adam Kirk, and Steve Sullivan.
\newblock High-quality streamable free-viewpoint video.
\newblock \emph{ACM Transactions on Graphics (TOG)}, 34\penalty0 (4):\penalty0
  69, 2015.

\bibitem[Dong et~al.(2016)Dong, Loy, He, and Tang]{Dong2016}
C.~Dong, C.~C. Loy, K.~He, and X.~Tang.
\newblock Image super-resolution using deep convolutional networks.
\newblock \emph{IEEE Trans. Pattern Anal. Machine Intelligence}, 38\penalty0
  (2):\penalty0 295--307, 2016.

\bibitem[Elhayek et~al.(2015)Elhayek, de~Aguiar, Jain, Tompson, Pishchulin,
  Andriluka, Bregler, Schiele, and Theobalt]{elhayek_efficient_2015}
Ahmed Elhayek, Edilson de~Aguiar, Arjun Jain, Jonathan Tompson, Leonid
  Pishchulin, Micha Andriluka, Chris Bregler, Bernt Schiele, and Christian
  Theobalt.
\newblock Efficient {ConvNet}-based marker-less motion capture in general
  scenes with a low number of cameras.
\newblock In \emph{Computer {Vision} and {Pattern} {Recognition} ({CVPR}), 2015
  {IEEE} {Conference} on}, pages 3810--3818, 2015.

\bibitem[Fattal(2007)]{Fattal2007}
R.~Fattal.
\newblock Image upsampling via imposed edge statistics.
\newblock In \emph{Proc. ACM SIGGRAPH}, 2007.

\bibitem[Gilbert et~al.(2018)Gilbert, Volino, Collomosse, and
  Hilton]{gilbert2018volumetric}
Andrew Gilbert, Marco Volino, John Collomosse, and Adrian Hilton.
\newblock Volumetric performance capture from minimal camera viewpoints.
\newblock In \emph{Proceedings of the European Conference on Computer Vision
  (ECCV)}, pages 566--581, 2018.

\bibitem[Glasner et~al.(2009)Glasner, Bagon, and Irani]{Glasner2009}
D.~Glasner, S.~Bagon, and M.~Irani.
\newblock Super-resolution from a single image.
\newblock In \emph{Proc. Intl. Conf. Computer Vision (ICCV)}, 2009.

\bibitem[Grauman et~al.(2003)Grauman, Shakhnarovich, and Darrell]{Grauman2003}
K.~Grauman, G.~Shakhnarovich, and T.~Darrell.
\newblock A bayesian approach to image-based visual hull reconstruction.
\newblock In \emph{Proc. CVPR}, 2003.

\bibitem[Hochreiter and Schmidhuber(1997)]{hochreiter1997LSTM}
Sepp Hochreiter and J{\"u}rgen Schmidhuber.
\newblock Long short-term memory.
\newblock In \emph{Neural computation}, volume~9, pages 1735--1780. MIT Press,
  1997.

\bibitem[Hossain and Little(2018)]{hossain2018exploiting}
Mir Rayat~Imtiaz Hossain and James~J Little.
\newblock Exploiting temporal information for 3d human pose estimation.
\newblock In \emph{European Conference on Computer Vision}, pages 69--86.
  Springer, 2018.

\bibitem[Huang et~al.(2017)Huang, Bogo, Classner, Kanazawa, Gehler, Akhter, and
  Black]{Huang3DV}
Yinghao Huang, Federica Bogo, Christoph Classner, Angjoo Kanazawa, Peter~V.
  Gehler, Ijaz Akhter, and J.~Black.
\newblock Towards accurate markerless human shape and pose estimation over
  time.
\newblock In \emph{3DV}, 2017.

\bibitem[Ionescu et~al.(2014)Ionescu, Papava, Olaru, and
  Sminchisescu]{h36m_pami}
Catalin Ionescu, Dragos Papava, Vlad Olaru, and Cristian Sminchisescu.
\newblock Human3.6m: Large scale datasets and predictive methods for 3d human
  sensing in natural environments.
\newblock \emph{IEEE Transactions on Pattern Analysis and Machine
  Intelligence}, 36\penalty0 (7):\penalty0 1325--1339, jul 2014.

\bibitem[Lan and Huttenlocher(2004)]{lan04}
X.~Lan and D.~Huttenlocher.
\newblock A unified spatio-temporal articulated model for tracking.
\newblock In \emph{Proc. Computer Vision and Pattern Recognition}, volume~1,
  pages 722--729, 2004.

\bibitem[Li et~al.(2015)Li, Zhang, and Chan]{li2015maximumH36m}
Sijin Li, Weichen Zhang, and Antoni~B Chan.
\newblock Maximum-margin structured learning with deep networks for 3d human
  pose estimation.
\newblock In \emph{Proceedings of the IEEE International Conference on Computer
  Vision}, pages 2848--2856, 2015.

\bibitem[Loper et~al.(2015)Loper, Mahmood, Romero, Pons-Moll, and
  Black]{loper2015SMPL}
Matthew Loper, Naureen Mahmood, Javier Romero, Gerard Pons-Moll, and Michael~J
  Black.
\newblock Smpl: A skinned multi-person linear model.
\newblock \emph{ACM Transactions on Graphics (TOG)}, 34\penalty0 (6):\penalty0
  248, 2015.

\bibitem[Malleson et~al.(2017)Malleson, Gilbert, Trumble, Collomosse, and
  Hilton]{Malleson3DV17}
C~Malleson, A~Gilbert, M~Trumble, J~Collomosse, and A~Hilton.
\newblock Real-time full-body motion capture from video and imus.
\newblock In \emph{3DV}, 2017.

\bibitem[Mude~Lin and Cheng(2017)]{lin2017CVPRRPSM}
Xiaodan Liang Keze~Wang Mude~Lin, Liang~Lin and Hui Cheng.
\newblock Recurrent 3d pose sequence machines.
\newblock In \emph{CVPR}, 2017.

\bibitem[Pavlakos et~al.(2017)Pavlakos, Zhou, Derpanis, and
  Daniilidis]{pavlakos2017volumetricCVPR}
Georgios Pavlakos, Xiaowei Zhou, Konstantinos~G Derpanis, and Kostas
  Daniilidis.
\newblock Coarse-to-fine volumetric prediction for single-image 3{D} human
  pose.
\newblock In \emph{CVPR}, 2017.

\bibitem[Qiu et~al.(2019)Qiu, Wang, Wang, Wang, and Zeng]{Qiu:iccv:2019}
Haibo Qiu, Chunyu Wang, Jingdong Wang, Naiyan Wang, and Wenjun Zeng.
\newblock Cross view fusion for 3d human pose estimation.
\newblock In \emph{Proceedings of the IEEE International Conference on Computer
  Vision}, 2019.

\bibitem[Ren and Collomosse(2012)]{ren12}
R~Ren and J~Collomosse.
\newblock Visual sentences for pose retrieval over low-resolution cross-media
  dance collections.
\newblock \emph{IEEE Transactions on Multimedia}, 2012.

\bibitem[Ren et~al.(2005)Ren, Berg, and Malik]{ren05}
X.~Ren, E.~Berg, and J.~Malik.
\newblock Recovering human body configurations using pairwise constraints
  between parts.
\newblock In \emph{Proc. Intl. Conf. on Computer Vision}, volume~1, pages
  824--831, 2005.

\bibitem[Rhodin et~al.(2016)Rhodin, Robertini, Casas, Richardt, Seidel, and
  Theobalt]{rhodin2016general}
Helge Rhodin, Nadia Robertini, Dan Casas, Christian Richardt, Hans-Peter
  Seidel, and Christian Theobalt.
\newblock General automatic human shape and motion capture using volumetric
  contour cues.
\newblock In \emph{European Conference on Computer Vision}, pages 509--526.
  Springer, 2016.

\bibitem[Rudin et~al.(1992)Rudin, Osher, and Fatemi]{tvexample}
L.~I. Rudin, S.~Osher, and E.~Fatemi.
\newblock Non-linear total variation based noise removal algorithms.
\newblock \emph{Physics D}, 60\penalty0 (1-4):\penalty0 259--268, 1992.

\bibitem[Shi et~al.(2016)Shi, Caballero, Huszar, Totz, Aitken, Bishop,
  Rueckert, and Wang]{Shi2016}
W.~Shi, J.~Caballero, F.~Huszar, J.~Totz, A.~Aitken, R.~Bishop, D.~Rueckert,
  and Z.~Wang.
\newblock Real-time single image and video super-resolution using an efficient
  sub-pixel convolutional neural network.
\newblock In \emph{Proc. Comp. Vision and Pattern Recognition (CVPR)}, 2016.

\bibitem[Srinivasan and Shi(2007)]{srinivasan07}
P.~Srinivasan and J.~Shi.
\newblock Bottom-up recognition and parsing of the human body.
\newblock In \emph{Proc. Computer Vision and Pattern Recognition}, pages 1--8,
  2007.

\bibitem[Starck et~al.(2009)Starck, Kilner, and Hilton]{starck2009FVVR}
Jonathan Starck, Joe Kilner, and Adrian Hilton.
\newblock A free-viewpoint video renderer.
\newblock \emph{Journal of Graphics, GPU, and Game Tools}, 14\penalty0
  (3):\penalty0 57--72, 2009.

\bibitem[Tan et~al.(2017)Tan, Budvytis, and Cipolla]{TanSMPLY2d3DBMVC17}
J~Tan, I~Budvytis, and R~Cipolla.
\newblock Indirect deep structured learning for 3d human body shape and pose
  prediction.
\newblock In \emph{BMVC}, 2017.

\bibitem[Tome et~al.(2017)Tome, Russell, and Agapito]{tome2017liftingH36m}
Denis Tome, Chris Russell, and Lourdes Agapito.
\newblock Lifting from the deep: Convolutional 3d pose estimation from a single
  image.
\newblock \emph{arXiv preprint arXiv:1701.00295}, 2017.

\bibitem[Toshev and Szegedy(2014)]{Toshev2014}
A.~Toshev and C.~Szegedy.
\newblock Deep pose: Human pose estimation via deep neural networks.
\newblock In \emph{Proc. CVPR}, 2014.

\bibitem[Trumble et~al.()Trumble, Gilbert, Malleson, Hilton, and
  Collomosse]{trumble_total_2017}
Matthew Trumble, Andrew Gilbert, Charles Malleson, Adrian Hilton, and John
  Collomosse.
\newblock Total capture: 3d human pose estimation fusing video and inertial
  sensors.
\newblock In \emph{Proceedings of 28th British Machine Vision Conference},
  pages 1--13.
\newblock URL \url{http://epubs.surrey.ac.uk/841740/}.

\bibitem[Trumble et~al.(2016)Trumble, Gilbert, Hilton, and
  John]{TrumbleCVMP2DConvNet}
Matthew Trumble, Andrew Gilbert, Adrian Hilton, and Collomosse John.
\newblock Deep convolutional networks for marker-less human pose estimation
  from multiple views.
\newblock In \emph{Proceedings of the 13th European Conference on Visual Media
  Production (CVMP 2016)}, CVMP 2016, 2016.

\bibitem[Trumble et~al.(2018)Trumble, Gilbert, Hilton, and
  Collomosse]{trumble:eccv:2018}
Matthew Trumble, Andrew Gilbert, Adrian Hilton, and John Collomosse.
\newblock Deep autoencoder for combined human pose estimation and body model
  upscaling.
\newblock In \emph{European Conference on Computer Vision (ECCV'18)}, 2018.

\bibitem[Varol et~al.(2018{\natexlab{a}})Varol, Ceylan, Russell, Yang, Yumer,
  Laptev, and Schmid]{JacksonECCV18}
G{\"{u}}l Varol, Duygu Ceylan, Bryan~C. Russell, Jimei Yang, Ersin Yumer, Ivan
  Laptev, and Cordelia Schmid.
\newblock Bodynet: Volumetric inference of 3d human body shapes.
\newblock In \emph{In ECCV'18}, 2018{\natexlab{a}}.

\bibitem[Varol et~al.(2018{\natexlab{b}})Varol, Ceylan, Russell, Yang, Yumer,
  Laptev, and Schmid]{VarolECCV18}
G{\"{u}}l Varol, Duygu Ceylan, Bryan~C. Russell, Jimei Yang, Ersin Yumer, Ivan
  Laptev, and Cordelia Schmid.
\newblock Bodynet: Volumetric inference of 3d human body shapes.
\newblock In \emph{In ECCV'18}, 2018{\natexlab{b}}.

\bibitem[von Marcard et~al.(2017)von Marcard, Rosenhahn, Black, and
  Pons-Moll]{SIP2017EG}
Timo von Marcard, Bodo Rosenhahn, Michael Black, and Gerard Pons-Moll.
\newblock Sparse inertial poser: Automatic 3d human pose estimation from sparse
  imus.
\newblock \emph{Computer Graphics Forum 36(2), Proceedings of the 38th Annual
  Conference of the European Association for Computer Graphics (Eurographics)},
  2017.

\bibitem[von Marcard et~al.(2018)von Marcard, Henschel, Black, Rosenhahn, and
  Pons-Moll]{PonsECCV18}
Timo von Marcard, Roberto Henschel, Michael~J Black, Bodo Rosenhahn, and Gerard
  Pons-Moll.
\newblock Recovering accurate 3d human pose in the wild using imus and a moving
  camera.
\newblock In \emph{Proceedings of the European Conference on Computer Vision
  (ECCV)}, pages 601--617, 2018.

\bibitem[Wang et~al.(2015)Wang, Liu, Yang, Han, and Huang]{Wang2015}
Z.~Wang, D.~Liu, J.~Yang, W.~Han, and T.~S. Huang.
\newblock Deep networks for image super-resolution with sparse prior.
\newblock In \emph{Proc. Intl. Conf. Computer Vision (ICCV)}, pages 370--378,
  2015.

\bibitem[Wei et~al.(2016)Wei, Ramakrishna, Kanade, and Sheikh]{wei2016cpm}
Shih-En Wei, Varun Ramakrishna, Takeo Kanade, and Yaser Sheikh.
\newblock Convolutional pose machines.
\newblock In \emph{CVPR}, 2016.

\bibitem[Xie et~al.(2012)Xie, Xu, and Chen]{Xie2012}
J.~Xie, L.~Xu, and E.~Chen.
\newblock Image denoising and inpainting with deep neural networks.
\newblock In \emph{Proc. Neural Inf. Processing Systems (NIPS)}, pages
  350--358, 2012.

\bibitem[Zheng et~al.(2019)Zheng, Yu, Wei, Dai, and Liu]{ZhengICCV19}
Zerong Zheng, Tao Yu, Yixuan Wei, Qionghai Dai, and Yebin Liu.
\newblock Deephuman: 3d human reconstruction from a single image.
\newblock In \emph{In ICCV'19}, 2019.

\bibitem[Zhu et~al.(2014)Zhu, Zhang, and Yuille]{Zhu2014}
Y.~Zhu, Y.~Zhang, and A.~L. Yuille.
\newblock Single image super-resolution using deformable patches.
\newblock In \emph{Proc. Comp. Vision and Pattern Recognition (CVPR)}, pages
  2917--2924, 2014.

\end{thebibliography}

   % name your BibTeX data base

\end{document}